\documentclass[sigconf]{aamas} 
\usepackage{balance} % for balancing columns on the final page
\usepackage{bm}
\usepackage{subcaption}
\usepackage{algorithm}
\usepackage[noend]{algpseudocode}

\setcopyright{ifaamas}
\acmConference[AAMAS '22]{Proc.\@ of the 21st International Conference
on Autonomous Agents and Multiagent Systems (AAMAS 2022)}{May 9--13, 2022}
{Online}{P.~Faliszewski, V.~Mascardi, C.~Pelachaud,
M.E.~Taylor (eds.)}
\copyrightyear{2022}
\acmYear{2022}
\acmDOI{}
\acmPrice{}
\acmISBN{}

\acmSubmissionID{???}

\title[AAMAS-2022 Formatting Instructions]{Automatic Calibration Framework of Agent-Based Models for Dynamic and Heterogeneous Parameters}
\subtitle{JAAMAS Track}

\author{Dongjun Kim}
\affiliation{
  \institution{KAIST}
  \city{Daejeon}
  \country{South Korea}}
\email{dongjoun57@kaist.ac.kr}

\author{Tae-Sub Yun}
\affiliation{
	\institution{KAIST}
	\city{Daejeon}
	\country{South Korea}}
\email{andy2402@kaist.ac.kr}

\author{Il-Chul Moon}
\affiliation{
	\institution{KAIST}
	\city{Daejeon}
	\country{South Korea}}
\email{icmoon@kaist.ac.kr}

\author{Jang Won Bae}
\affiliation{
	\institution{Korea University of Technology and Education}
	\city{Cheonan}
	\country{South Korea}}
\email{jangwon_bae@koreatech.ac.kr}

\begin{abstract}
Agent-based models (ABMs) highlight the importance of simulation validation, such as qualitative face validation and quantitative empirical validation. In particular, we focused on quantitative validation by adjusting simulation input parameters of the ABM. This study introduces an \textit{automatic calibration framework} that combines the suggested dynamic and heterogeneous calibration methods. Specifically, the dynamic calibration fits the simulation results to the real-world data by automatically capturing suitable simulation time to adjust the simulation parameters. Meanwhile, the heterogeneous calibration reduces the distributional discrepancy between individuals in the simulation and the real world by adjusting agent related parameters cluster-wisely. 
\end{abstract}


\newcommand{\BibTeX}{\rm B\kern-.05em{\sc i\kern-.025em b}\kern-.08em\TeX}

\begin{document}

%%% The following commands remove the headers in your paper. For final 
%%% papers, these will be inserted during the pagination process.

\pagestyle{fancy}
\fancyhead{}

%%% The next command prints the information defined in the preamble.

\maketitle 

%%%%%%%%%%%%%%%%%%%%%%%%%%%%%%%%%%%%%%%%%%%%%%%%%%%%%%%%%%%%%%%%%%%%%%%%

\section{Introduction}

Recently, enhanced computing power has motivated the construction of Agent-Based Models (ABMs) in a highly complex manner, and their efficacy has been expanded to various domains, such as market modeling \cite{bonabeau2002agent}, traffic management \cite{naiem2010agent}, and urban planning \cite{hosseinali2013agent}. According to this extended applicability, the accuracy of the ABM compared to the target real-world is also in demand. Therefore, validation of the ABM becomes essential \cite{beisbart2018computer}. 

ABM naturally diverges from the real-world because of temporal discrepancies and agent heterogeneity. To fit the simulation to the real-world, we introduce two distinctive calibration methods regulating each diverging source: one is the dynamic calibration that adjusts the input parameters to vary over the simulation time to improve the temporal fitness. However, it could be computationally prohibitive to optimize dynamically varying parameters if we tune dynamic parameters at every time tick; therefore, we consider the regime as the smallest unit in parameter diversification. The regime is an object to be optimized via the Hidden Markov Model (HMM) \cite{bishop2006pattern} for every iteration. The other is the heterogeneous calibration that fits the simulation to the real-world by diversifying parameters and optimizing these diversified parameters, which are likely to differ by agents. In this case, the agent cluster becomes the smallest unit of parameter diversification in heterogeneous calibration to limit the computational bottleneck. We obtain agent clusters by applying a Gaussian Mixture Model (GMM) \cite{bishop2006pattern} to the latent embeddings extracted by the Variational Auto-Encoder (VAE) \cite{kingma2013auto}.

As a generalization of two methods, we introduce a \textit{calibration framework} of the ABM (Algorithm \ref{alg:algorithm}) by interchangeably adjusting dynamic parameters with $C_{dyn}$ consecutive iterations and heterogeneous parameters with $C_{het}$ consecutive iterations. Notably, the \textit{calibration framework} reduces to the dynamic (or heterogeneous) calibration if $C_{het}=0$ (or $C_{dyn}=0$). In experiments, we established that each calibration method and their joint \textit{calibration framework} significantly improve the simulation fitness to the real-world. 

\vspace{-0.05cm}
\begin{algorithm}[h]
	\caption{Calibration Framework of ABM}
	\label{alg:algorithm}
	\begin{algorithmic}[1]
		\State Select the dynamic and heterogeneous parameters
		\State Obtain agent clusters via the GMM and VAE
		\Repeat
		\For{$i=1$ to $C_{dyn}$} \do \hfill\Comment{Dynamic Calibration}
		
		\State Detect temporal regimes via HMM
		\State Optimize dynamic parameters
		\EndFor
		\For{$j=1$ to $C_{het}$} \do \hfill\Comment{Heterogeneous Calibration}
		
		\State Optimize heterogeneous parameters
		\EndFor
		\Until{converged}
	\end{algorithmic}
\end{algorithm}
\vspace{-0.2cm}

\section{Details in Two Calibration Methods}

\textbf{Dynamic Calibration} The dynamic calibration is a particle-based approach \cite{sisson2007sequential} that iteratively updates the proposal distribution, which is $q(\bm{\theta}\vert\mathbf{x}_{o};\bm{\phi})$ parametrized by $\bm{\phi}$, where $\mathbf{x}_{o}$ is the single-shot real-world observation and $\bm{\theta}$ is the calibration target parameters. The next candidate particles are sampled from this proposal distribution, and the calibration at the end estimates the optimal set of parameters that best fits the simulation to the real-world after calibration iterations. Based on the Bayes rule, the proposal distribution satisfies $q(\bm{\theta}\vert\mathbf{x}_{0};\bm{\phi})\propto q(\bm{\theta}\vert\bm{\phi})p(\mathbf{x}_{o}\vert\bm{\theta})$, where $q(\bm{\theta}\vert\bm{\phi})$ and $p(\mathbf{x}_{o}\vert\bm{\theta})$ are the building blocks to model the proposal distribution. According to \citet{wood2010statistical}, we estimate $p(\mathbf{x}_{o}\vert\bm{\theta})$ as an empirical Gaussian distribution estimated from 10 simulation replications. Additionally, we model $q(\bm{\theta}\vert\bm{\phi})$ as a product of Beta distributions $q(\bm{\theta}\vert\bm{\phi})=\Pi_{r=1}^{R}\text{Beta}(\theta_{r}\vert\phi_{r})$, where $\theta_{r}$ is the parameter value of the $r$-th regime, and $\phi_{r}:=\{\alpha_{r},\beta_{r}\}$ is the shape coefficients of the Beta distribution of the $r$-th regime. We update the proposal distribution by maximizing $q(\bm{\theta}\vert\mathbf{x}_{o};\bm{\phi})$ with respect to $\bm{\phi}$ each iteration. 

\textbf{Heterogeneous Calibration} The heterogeneous calibration works by Bayesian optimization. The Bayesian optimization \cite{frazier2018tutorial} is applied to a surrogate model of the fitness function estimated by the Gaussian process \cite{tresidder2012acceleration}. The approach using a surrogate model has been introduced previously in many disciplines. One distinctive point from previous research is that we separate agents by clusters and assign diverse parameter values to the divided clusters. Notably, the response curve of the ABM is sometimes non-differentiable at branch points, where the emergent behavior does not arise \textit{unless} the parameter value reaches to such points. Because the most prominent Expected Improvement (EI) acquisition function sometimes fails to converge to the global minimum when the response curve is non-differentiable \cite{bodin2019modulated}, our strategy involves mixing various acquisition functions \cite{hoffman2011portfolio} to optimize the heterogeneous parameters. We propose the next set of candidate parameters randomly selected from 1) random sample, 2) max argument of predictive variance (exploration), 3) min argument of predictive mean (exploitation), and 4) max argument of weighted Expected Improvement \cite{sobester2005design}. 

\section{Experiments}

\begin{figure}[t]
\centering
\begin{subfigure}{0.48\linewidth}
		\centering
    \includegraphics[width=1.1\linewidth]{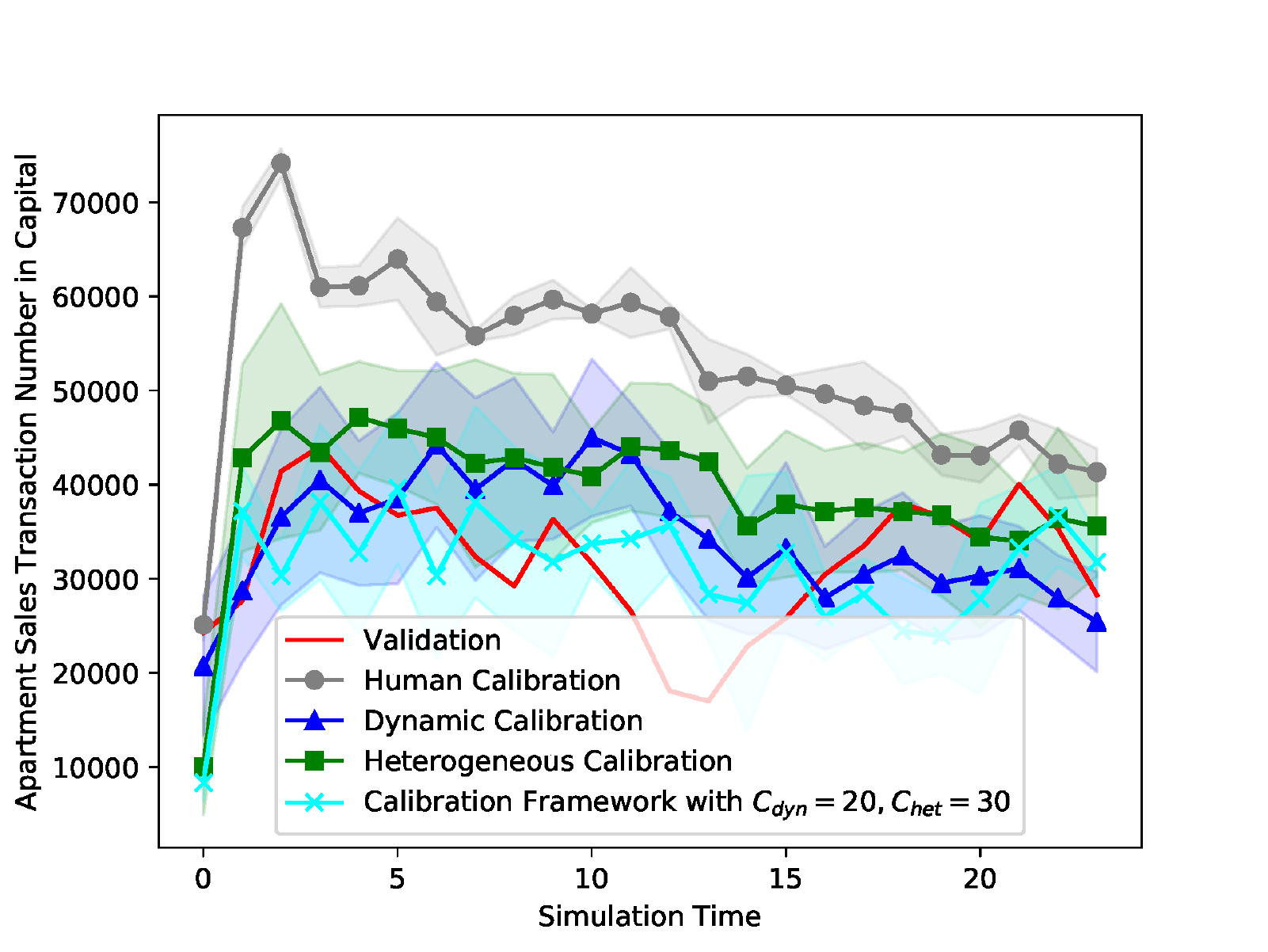}
    \subcaption{Calibrated Results}
	\end{subfigure}
	\begin{subfigure}{0.48\linewidth}
		\centering
    \includegraphics[width=1.1\linewidth]{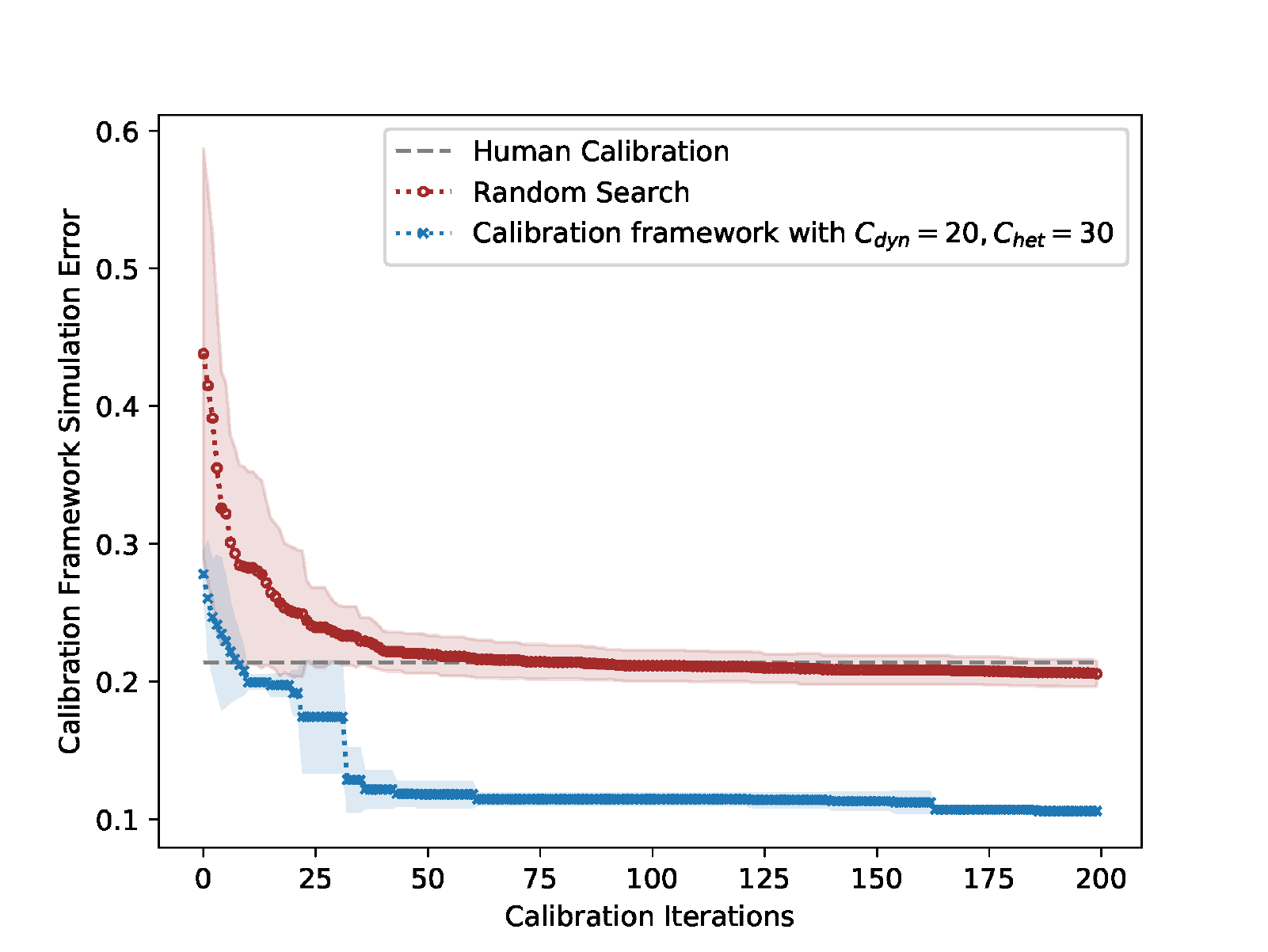}
    \subcaption{Error by Iterations}
	\end{subfigure}
	\vskip -0.05in
	\caption{(a) compares the manual calibration with the suggested methods. (b) presents MAPE by calibration iterations.}
	\label{fig:fig1}
	\vskip -0.3in
\end{figure}

We tested our calibration methods with the real estate market ABM of South Korea \cite{yun2020housing}. The real-world housing market consists of various types of agents and is significantly affected by economic trends; therefore, the market is a favorable scenario for testing our calibration methods. There are three dynamic parameters: \textit{Market-Participation-Rate}, \textit{Market-Price-Increase-Rate}, and \textit{Market-Price-Decrease-Rate}, as well as two heterogeneous parameters: \textit{Willing-to-Pay} and \textit{Purchase-Rate}. These are unobservable parameters that determine the underlying demand and supply curve of the model, so these are the representative parameters to calibrate.

Figure \ref{fig:fig1}-(a) compares the observation with 1) manual human calibration, 2) dynamic calibration, 3) heterogeneous calibration, and 4) combined calibration. Both of the suggested calibration methods significantly improve the human manual calibration. For Mean Absolute Percentage Error (MAPE), the human calibration is 0.765, whereas the MAPE is reduced to 0.281 in the dynamic calibration, and 0.232 in the heterogeneous calibration. In addition, we obtain a simulation that is best suited to the observation by combining two calibration methods with a MAPE of 0.219. Figure \ref{fig:fig1}-(b) demonstrates that the suggested \textit{calibration framework} outperforms to random search and human calibration. 

\begin{figure}[t]
	\centering
	\begin{subfigure}{0.48\linewidth}
		\centering
		\includegraphics[width=1.1\linewidth]{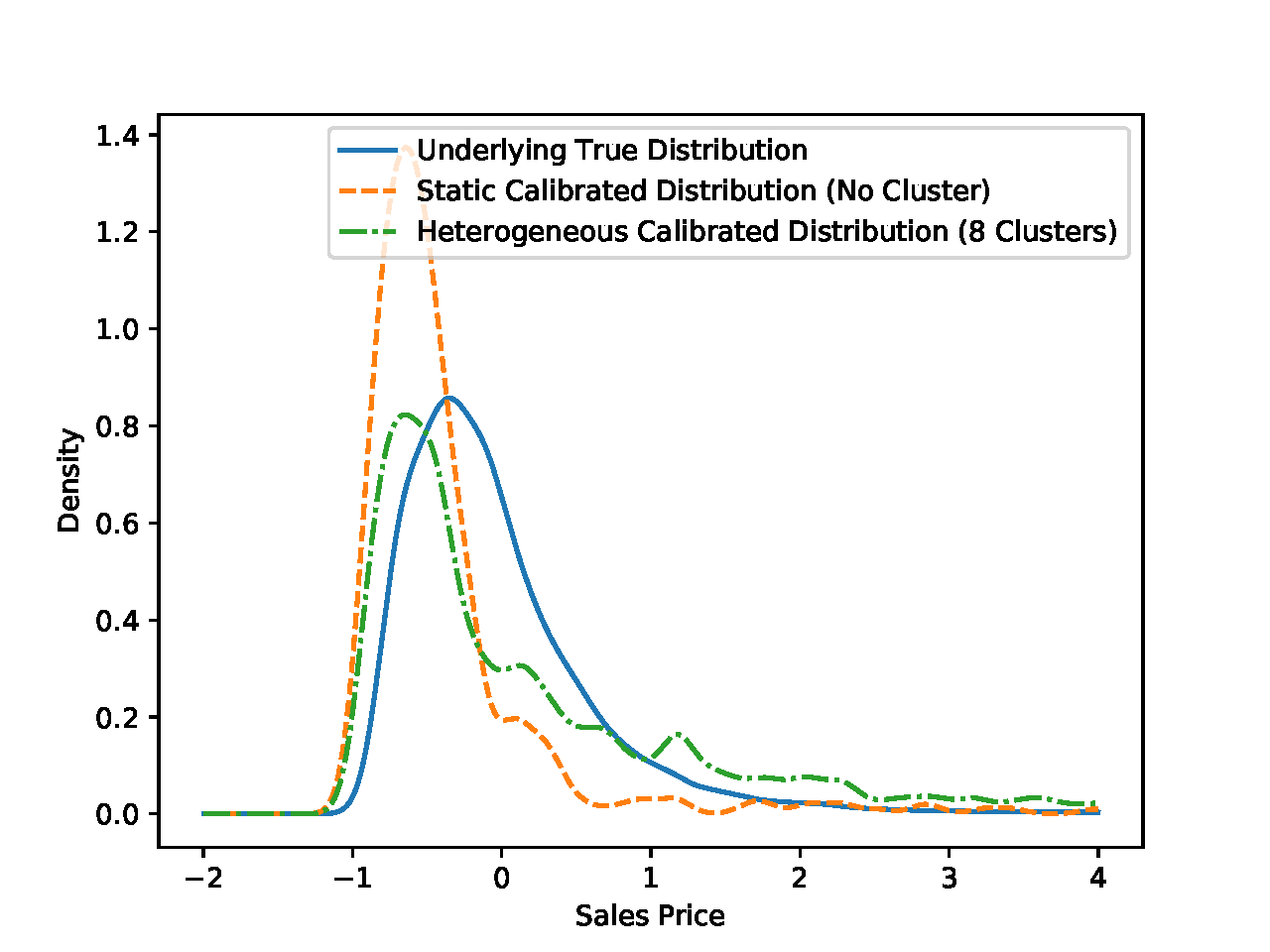}
		\subcaption{Micro-level Distributions}
	\end{subfigure}
	\begin{subfigure}{0.48\linewidth}
		\centering
		\includegraphics[width=1.1\linewidth]{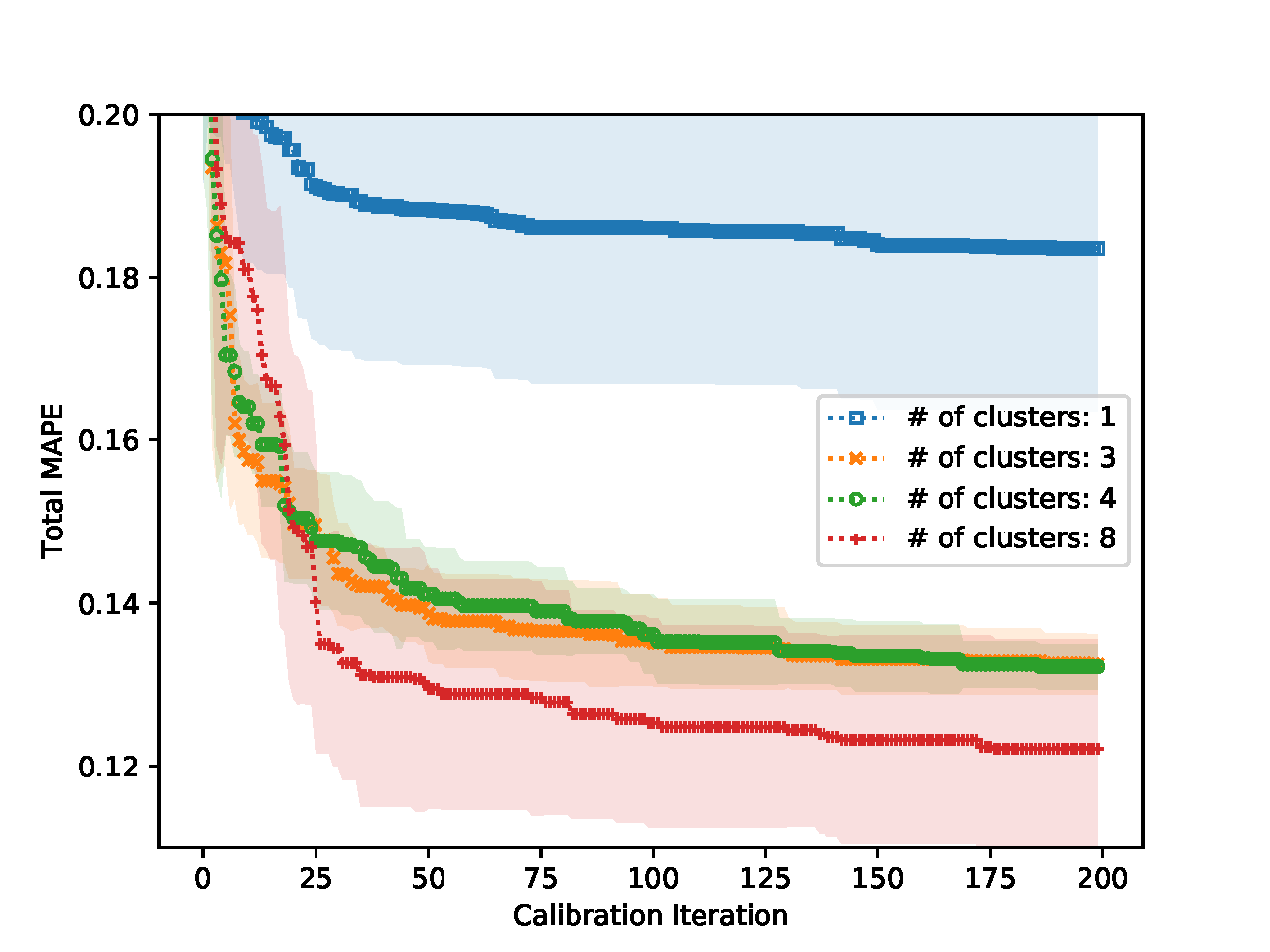}
		\subcaption{Error by $\#$Clusters}
	\end{subfigure}
	\vskip -0.05in
	\caption{(a) shows that the agent heterogeneity is largely fitted by heterogeneous calibration. (b) illustrates that fitting the distributional divergence is beneficial on validation.}
	\label{fig:fig2}
	\vskip -0.2in
\end{figure}

\begin{figure}[t]
\centering
	\begin{subfigure}{0.48\linewidth}
		\centering
    \includegraphics[width=1.1\linewidth]{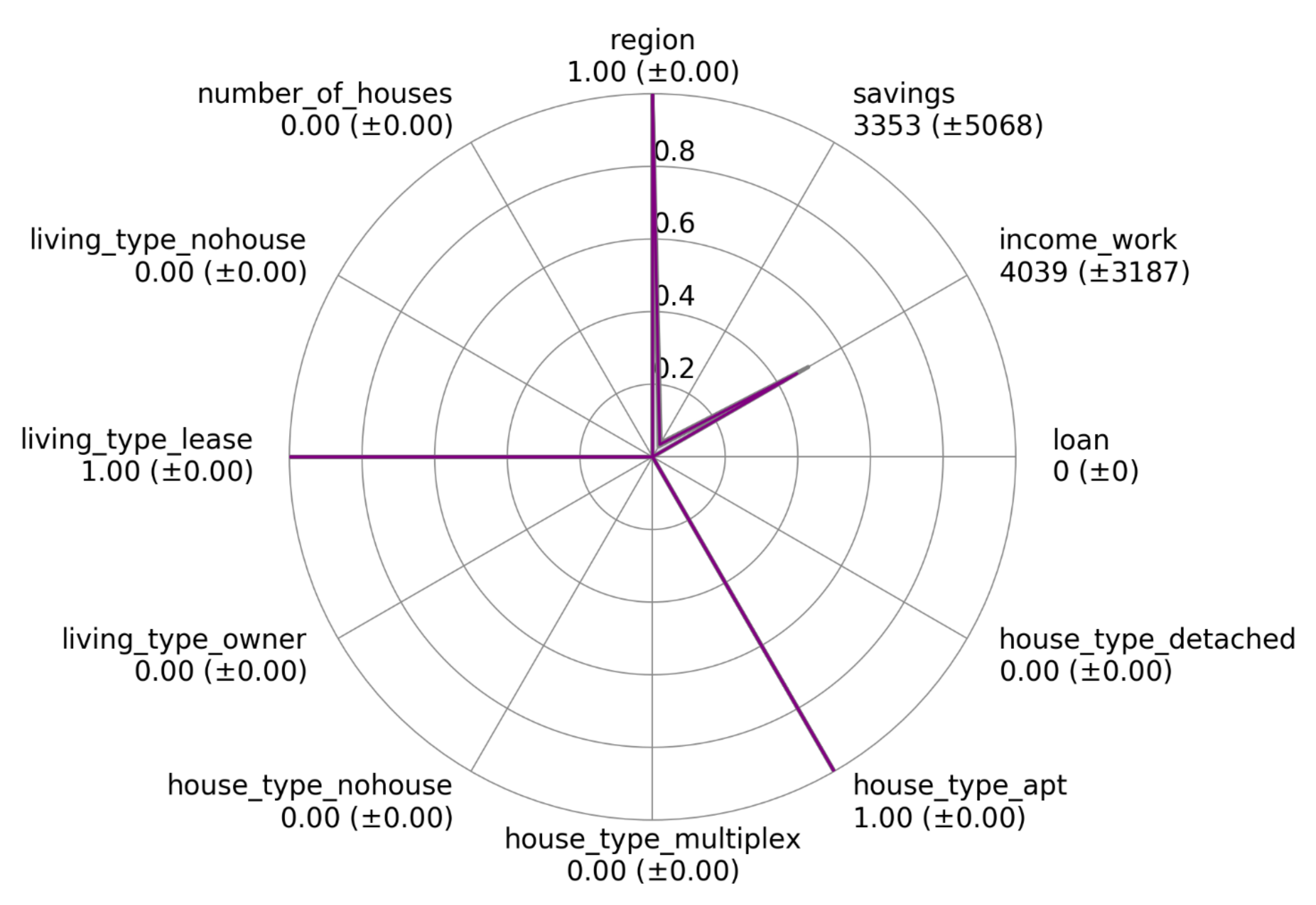}
	\end{subfigure}
	\begin{subfigure}{0.48\linewidth}
		\centering
    \includegraphics[width=1.1\linewidth]{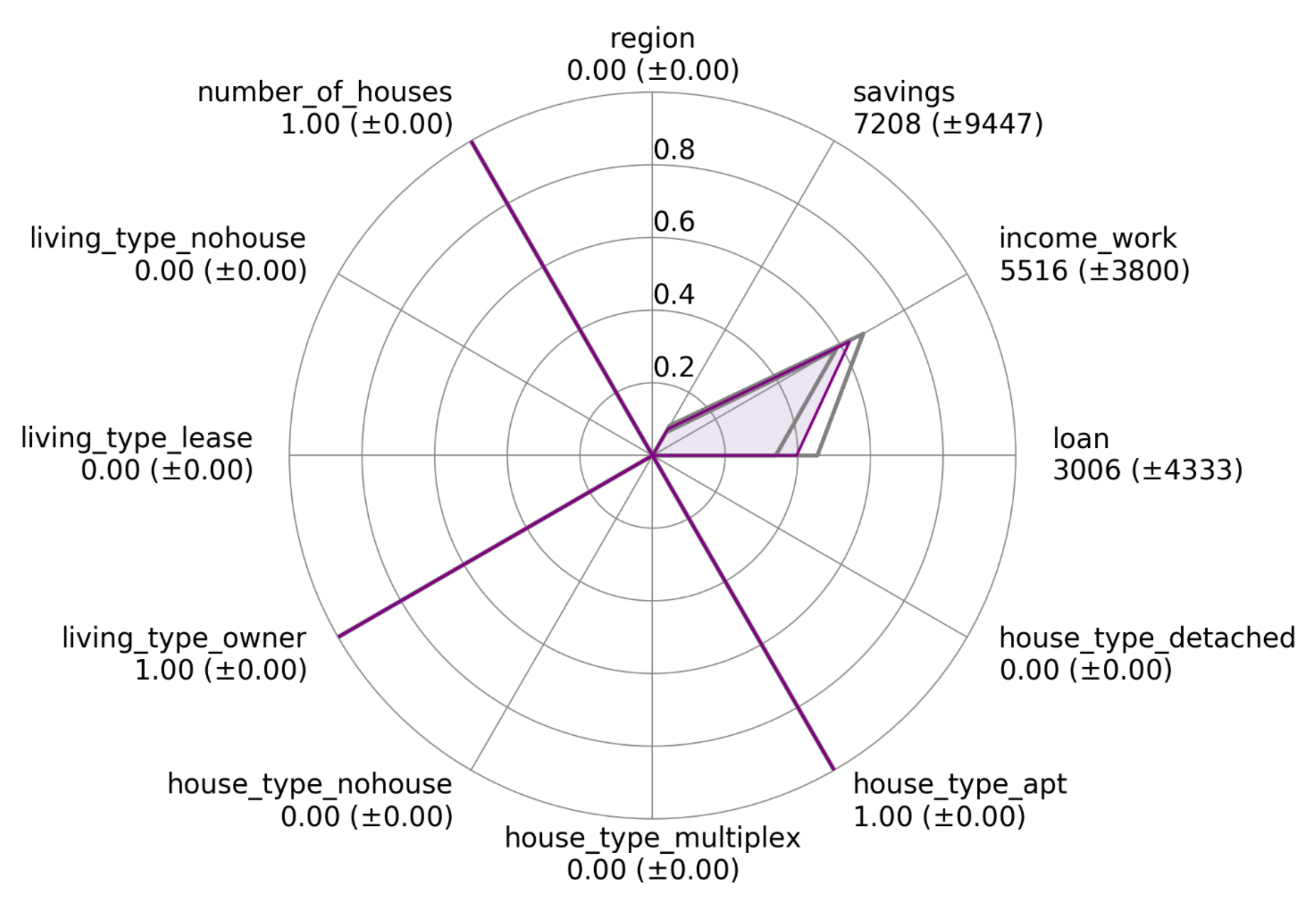}
	\end{subfigure}
	\caption{Two example agent-clusters.}
	\label{fig:fig3}
		\vskip -0.3in
\end{figure}

Figure \ref{fig:fig2}-(a) presents the validity of heterogeneous calibration by showing that undifferentiated parameter is too rigid to fit the true distribution. Conversely, owing to the expanded degree of freedom, the heterogenous calibration not only improves the targeted summary statistics, but also fits the overall distribution as in Figure \ref{fig:fig2}-(b). In Figure \ref{fig:fig3}, agents in the left/right clusters live in rental/own houses, and this difference leads the optimal \textit{Willing-to-Pay} to be 0.9/0.3 for left/right clusters, respectively. This indicates that agents in the left cluster without their own house are more willing to buy a new house in the near future.

\section{Conclusion}

This study proposes an automatic \textit{calibration framework} of the ABM that generalizes both dynamic and heterogeneous calibrations, which discovered a well-calibrated parameter sets in experiments.

\begin{acks}
This research was supported by Development of City Interior Digital Twin Technology to establish Scientific Policy through the Institute for Information $\&$ communication Technology Planning $\&$ evaluation(IITP) funded by the Ministry of Science and ICT(2018-0-00225).
\end{acks}

%%%%%%%%%%%%%%%%%%%%%%%%%%%%%%%%%%%%%%%%%%%%%%%%%%%%%%%%%%%%%%%%%%%%%%%%

%%% The acknowledgments section is defined using the "acks" environment
%%% (rather than an unnumbered section). The use of this environment 
%%% ensures the proper identification of the section in the article 
%%% metadata as well as the consistent spelling of the heading.

%%%%%%%%%%%%%%%%%%%%%%%%%%%%%%%%%%%%%%%%%%%%%%%%%%%%%%%%%%%%%%%%%%%%%%%%

%%% The next two lines define, first, the bibliography style to be 
%%% applied, and, second, the bibliography file to be used.

\bibliographystyle{ACM-Reference-Format} 
\bibliography{reference}

%%%%%%%%%%%%%%%%%%%%%%%%%%%%%%%%%%%%%%%%%%%%%%%%%%%%%%%%%%%%%%%%%%%%%%%%

\end{document}